\documentclass{article}

\usepackage[final]{neurips_2019}

\usepackage[utf8]{inputenc} 
\usepackage[T1]{fontenc}    
\usepackage{hyperref}       
\usepackage{url}            
\usepackage{booktabs}       
\usepackage{amsmath}
\usepackage{amsthm}
\usepackage{amsfonts}       
\usepackage{nicefrac}       
\usepackage{microtype}      
\usepackage{tikz}
\usepackage{cancel}
\usepackage{subcaption}
\usepackage[mark=***]{sectionbreak}
\usetikzlibrary{arrows}

\theoremstyle{definition}

\theoremstyle{definition}
\newtheorem{definition}{Definition}[]

\usepackage[acronym,smallcaps,nowarn,section,nogroupskip,nonumberlist]{glossaries}
\newacronym{mab}{MAB}{Multihead Attention Block}
\newacronym{sab}{SAB}{Self-Attention Block}
\newacronym{isab}{ISAB}{Induced Self-Attention Block}
\newacronym{mpnn}{MPNN}{Message Passing Neural Network}
\newacronym{gnf}{GNF}{Graph Normalizing Flow}
\newacronym{ge-vae}{GE-VAE}{Graph Embedding VAE}
\newacronym{mmd}{MMD}{Maximum Mean Discrepancy}

\title{Graph Embedding VAE: A Permutation\\ Invariant Model of Graph Structure}

\author{
    Tony Duan\\
    \texttt{tonyduan@cs.stanford.edu}\thanks{Work done as an intern at AITRICS.}
    \And
    Juho Lee\\
    \texttt{juho@aitrics.com}
}

\begin{document}

\maketitle

\begin{abstract}
    Generative models of graph structure have applications in biology and social sciences. The state of the art is GraphRNN, which decomposes the graph generation process into a series of sequential steps. While effective for modest sizes, it loses its permutation invariance for larger graphs. Instead, we present a permutation invariant latent-variable generative model relying on graph embeddings to encode structure. Using tools from the random graph literature, our model is highly scalable to large graphs with likelihood evaluation and generation in $O(|V| + |E|)$.
\end{abstract}

\section{Method}

We focus on learning a generative model of un-directed graph structure without node labels. Let $G = (V,E)$ denote a graph represented by a symmetric adjacency matrix $\mathbf{A} \in \{0,1\}^{|V|\times |V|}$. Note that we are interested in the \emph{inductive} (across graphs) setting rather than transductive (single graph).

\paragraph{Permutation invariance.}

We begin with a few useful definitions \citep{zaheer_deep_2017}.
\begin{definition}
    A function $f: \mathcal{X}^n \rightarrow \mathcal{Y}$ is \emph{permutation-invariant} if and only if it satisfies 
    $f(\pi x) = f(x)$ for any permutation $\pi \in S_n$, the set of permutations of indices $\{1,\hdots,n\}$.
\end{definition}
\begin{definition}
    A function $f: \mathcal{X}^n \rightarrow \mathcal{Y}^n$ is \emph{permutation-equivariant} if and only if it satisfies 
    $f(\pi x) = \pi f(x)$ for any permutation $\pi \in S_n$, the set of permutations of indices $\{1,\hdots, n\}$.
\end{definition}

Within the context of graphs, the \gls*{mpnn} \citep{gilmer_neural_2017} is the most popular permutation-equivariant model. In its simplest form, a \gls*{mpnn} acts on a set of node features $\mathbf{X}$ given a fixed adjacency matrix $\mathbf{A}$ by layers of message passing, described by
\begin{equation*}
    \mathrm{MPNNLayer}(\mathbf{X}; \mathbf{A}) = \sigma(\mathbf{A}\mathbf{X}\mathbf{W}), \quad\quad \text{where }\sigma \text{ is a non-linearity}.
\end{equation*}
Another permutation-equivariant model is the Set Transformer, which is composed of layers called the \gls*{isab}. Based on multihead attention~\citep{vaswani_attention_2017}, an \gls*{isab}
computes the pairwise interactions between the $n$ elements in $\mathbf{X}$. 
\begin{equation*}
\mathrm{ISAB}(\mathbf{X}) = \mathrm{MultiheadAttention}(\mathbf{X}, 
\mathrm{MultiheadAttention}(\mathbf{I}, \mathbf{X})),
\end{equation*}
where $\mathbf{I}$ is a set of $m$ trained inducing points.
Instead of computing self-attention directly on $\mathbf{X}$ requiring $O(n^2)$ time complexity, \gls*{isab} indirectly compares the elements in $\mathbf{X}$
via the reference points $\mathbf{I}$, thus reducing the time-complexity to $O(nm)$. 
It is worth noting that an \gls*{isab} is a special case of a \gls*{mpnn} with a fully connected adjacency matrix $\mathbf{A}$. 

\paragraph{Joint permutation invariance.}

To satisfy permutation invariance with respect to arbitrary node re-orderings, we want to learn a likelihood model $p_\theta$ such that:
\begin{equation*}
p_\theta(\mathbf{PAP}^\top) = p_\theta(\mathbf{A}), \text{for all permutation matrices }\mathbf{P}.
\end{equation*}
Note that this is a different type of symmetry than the permutation equivariance we've already described, because the rows \emph{and} columns of $\mathbf{A}$ need to be re-ordered when the permutation is applied. Following \citet{bloem-reddy_probabilistic_2019}, we call this type of symmetry \emph{joint invariance}.

\begin{definition}
    A function $f: \mathcal{X}^{n\times n} \rightarrow \mathcal{Y}$ is \emph{jointly permutation-invariant} if and only if it satisfies $f(\pi x \pi^\top) =  f(x)$ for any permutation $\pi \in S_n$, the set of permutations of indices $\{1,\hdots,n\}$.
\end{definition}
\begin{definition}
    A function $f: \mathcal{X}^{n\times n} \rightarrow \mathcal{Y}^n$ is \emph{jointly permutation-equivariant} if and only if it satisfies $f(\pi x \pi^\top) =  \pi f(x)$ for any permutation $\pi \in S_n$, the set of permutations of indices $\{1,\hdots,n\}$.
\end{definition}



\paragraph{Latent-variable models.}

\begin{figure}[t]
\centering
\begin{tikzpicture}
\tikzset{vertex/.style = {shape=circle,draw,minimum size=2em}}
\node[vertex] (A) at  (0,0) {$\mathbf{A}$};
\node[vertex] (X) at  (2,0) {$\mathbf{Z}_0$};
\node[vertex] (Z) at  (4,0) {$\mathbf{Z}$};
\draw[->] (A) to [bend right=1] node[below] {$\mathrm{Embed}$} (X);
\draw[<->] (Z) to node[below] {$\mathbf{f}_\phi$} (X);
\draw[->] (X) to [bend right] node[above] {$\mathrm{Reconstruct}$} (A);
\end{tikzpicture}
\caption{Graphical model summarizing our approach. We use a permutation-equivariant graph embedding $\mathbf{A} \rightarrow \mathbf{Z}_0$ to encode adjacency matrices into latent space, then apply a normalizing flow $\mathbf{Z}_0 \leftrightarrow \mathbf{Z}$. We rely on an approximate posterior for reconstruction $\mathbf{Z}_0 \rightarrow \mathbf{A}$.}
\label{fig:pgm}
\end{figure}
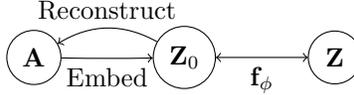

Instead of learning the adjacency matrix $p_\theta(\mathbf{A})$ directly, we introduce a latent variable $\mathbf{Z}$. Following the standard formulation of variational auto-encoders \citep{kingma_auto-encoding_2014}, we introduce a variational approximation to model the intractable posterior.
\begin{equation*}
p_\theta(\mathbf{A}) = \int_{\mathbf{Z}} p(\mathbf{Z}) p_\theta(\mathbf{A}|\mathbf{Z})\quad\quad q_\phi(\mathbf{Z}|\mathbf{A}) \approx p_\theta(\mathbf{Z}|\mathbf{A})
\end{equation*}
This puts our method in the same line of work as VGAE \citep{kipf_variational_2016} and Graphite \citep{grover_graphite:_2019}. Both are latent-variable models with permutation-equivariant decoders $p_\theta(\mathbf{A}|\mathbf{Z},\mathbf{X})$ and permutation-equivariant encoders $q_\phi(\mathbf{Z}|\mathbf{A},\mathbf{X})$, instantiated as message-passing neural networks \citep{gilmer_neural_2017}. However, these prior works rely on the availability of node features $\mathbf{X}$ for message passing. In the absence of node features an arbitrary ordering of nodes is used to learn initial embeddings (by setting $\mathbf{X} = \mathbf{I}_{n}$), resulting in the loss of permutation-equivariance.

\paragraph{Graph embeddings.} 

How do we break symmetries between nodes in the absence of node features? We explore use of a graph embedding to encode structure of the graph. Formally, we employ a function $\mathrm{Embed} : \{ 0, 1\}^{|V|\times |V|} \rightarrow \mathbb{R}^{|V|\times P}$ that satisfies joint permutation-equivariance. The textbook example of a graph embedding method is the Laplacian Eigenmap \citep{belkin_laplacian_2003,verma_hunt_2017}, defined via the eigendecomposition of the Laplacian matrix.
\begin{equation*}
    \mathrm{Embed}(\mathbf{A}) = \mathbf\Phi^\top, \quad\quad \mathbf{D - A} = \mathbf\Phi \mathbf\Lambda \mathbf\Phi^\top.
\end{equation*}
The Laplacian Eigenmap is our canonical example of an embedding method, though in experiments we investigate Locally Linear Embeddings as well \citep{roweis_nonlinear_2000}. More recently developed deep learning embeddings built on stochastic random walks could theoretically be employed as well \citep{perozzi_deepwalk:_2014,grover_node2vec:_2016,abu-el-haija_watch_2018}. However, we note that such methods are typically invariant to permutations of the embedding dimensions, resulting in a different type of symmetry, so we leave their investigation to future work.

\paragraph{Encoder.} We begin our variational posterior with a graph embedding, then apply a normalizing flow to improve expressivity \citep{rezende_variational_2015}. Letting $\mathbf{f}_\phi$ denote a differentiable invertible transformation (potentially composed of a chain of simpler such tranformations), we have
\begin{align*}
    \mathbf{Z}_0|\mathbf{A} &\sim \mathrm{Normal}(\mathrm{Embed}(\mathbf{A}), \sigma^2 \mathbf{I})\quad\quad \mathbf{Z} = \mathbf{f}_\phi(\mathbf{Z}_0)\\
    \log q_\phi(\mathbf{Z}|\mathbf{A}) &= \log q(\mathbf{Z}_0|\mathbf{A}) - \log\left|\det \frac{\partial \mathbf{f}_\phi}{\partial \mathbf{Z}_0}\right|.
\end{align*}
We parameterize $\mathbf{f}_\phi$ as a Neural Spline Flow over coupling layers \citep{durkan_neural_2019} for adequate expressivity. Coupling and $1 \times 1$ convolutions are performed over the $P$ dimensions of the embeddings. To ensure permutation equivariance while allowing dependencies between nodes, the splines for each coupling layer are parameterized by a stack of \gls*{isab}s. We note that stacking self-attention layers over the node embeddings results in potentially complex interactions between nodes that have typically been captured via message-passing neural networks. However, the use of \gls*{isab}s requires only $O(|V|m)$ complexity instead of $O(|V|^2)$ complexity, where $m$ is the number of inducing points.

\paragraph{Decoder.} Our decoder applies the inverse flow $\mathbf{f}_\phi^{-1}$, an \gls*{isab}, then a Bernoulli-Exponential link.
\begin{align*}
    \mathbf{Z} &\sim \mathrm{Normal}(\mathbf{0}, \mathbf{I}) \quad\quad \mathbf{Z}^\ast = \mathrm{ISABStack}(\mathbf{f}_\phi^{-1}(\mathbf{Z}))\\
    \mathbf{A} | \mathbf{Z}^\ast &\sim \mathrm{Bernoulli}\textnormal{-}\mathrm{Exponential}(\mathbf{Z}^\ast\mathbf{Z}^{\ast^\top})
\end{align*}
The Bernoulli-Exponential link \citep{zhou_infinite_2015,caron_bayesian_2012} is defined by augmenting the model with truncated Exponential random variables $\mathbf{M} \in [0,1]^{|V| \times |V|}$ corresponding to entries of the adjacency matrix $\mathbf{A}$. Letting $\mathbf{z}_i^\ast, \mathbf{z}_j^\ast$ denote rows of $\mathbf{Z}^\ast$ corresponding to the $i$-th and $j$-th nodes, 
\begin{equation*}
    a_{i,j} | \mathbf{z}^\ast_i, \mathbf{z}^\ast_j \sim \mathrm{Bernoulli}(m_{i,j} < 1) \quad\quad m_{i,j} | \mathbf{z}^\ast_i, \mathbf{z}^\ast_j \sim \mathrm{Exponential}(\mathbf{z}_i^{\ast^\top} \mathbf{z}^\ast_{j})
\end{equation*}

The joint log-likelihood can then be expressed as
\begin{equation*}
    \log p_\theta(\mathbf{A}, \mathbf{M}|\mathbf{Z}^\ast) = \sum_{(i,j)\in E}\left( \log(\mathbf{z}_i{\ast^\top} \mathbf{z}_j^\ast)  - \mathbf{z}_i^{\ast^\top} \mathbf{z}^\ast_j (m_{i,j} - 1)\right) - \frac{1}{2} \left( (\sum_{i=1}^n \mathbf{z}^\ast_i)^\top (\sum_{i=1}^n \mathbf{z}^\ast_i)- \sum_{i=1}^n ||\mathbf{z}^\ast_i||^2_2 \right).
\end{equation*}
The advantage of the Bernoulli-Exponential link function over a traditional (ex. logistic) link function is scalability; the joint log-likelihood can be can be calculated in $O(|E| + |V|)$ instead of $O(|V|^2)$. We sample from the analytic posterior for inference, noting that we only need to sample the $|E|$ auxiliary variables where $a_{i,j} = 1$ ($\delta_1$ below denotes the Dirac delta function centered at $1$). Since the auxiliary random variables are continuous the re-parameterization trick can be used.
\begin{equation*}
    q(\mathbf{M}|\mathbf{A},\mathbf{Z}^\ast) = \prod_{i < j} q(m_{i,j}|a_{i,j},\mathbf{z}^\ast_i,\mathbf{z}^\ast_j)\quad\quad q(m_{i,j}|a_{i,j}) = (1-a_{i,j})\delta_1 + a_{i,j} p(m_{i,j}|\mathbf{z}_i^\ast, \mathbf{z}_j^\ast)
\end{equation*}

\paragraph{Summary.} The high-level idea is summarized in Figure \ref{fig:pgm}. We fit by optimizing the ELBO,
\begin{align*}
    \log p_\theta(\mathbf{A}) & \geq \mathbb{E}_{q_\phi(\mathbf{Z},\mathbf{M}|\mathbf{A})}[\log p_\theta(\mathbf{A},\mathbf{Z},\mathbf{M}) - \log q_\phi(\mathbf{Z},\mathbf{M}|\mathbf{A})]\\
    & = \mathbb{E}_{q(\mathbf{Z}_0,\mathbf{M}|\mathbf{A})}[\log p(\mathbf{Z}) + \log p_\theta(\mathbf{A},\mathbf{M}| \mathbf{Z}) - \log q_\phi(\mathbf{Z}|\mathbf{A}) - \log q(\mathbf{M}|\mathbf{A},\mathbf{Z})]\\
    & = \mathbb{E}_{q(\mathbf{Z}_0,\mathbf{M}|\mathbf{A})}\left[\log p(\mathbf{Z}) + \log p_\theta(\mathbf{A},\mathbf{M}|\mathbf{Z}_0) - \log q(\mathbf{Z}_0|\mathbf{A}) -\log q(\mathbf{M}|\mathbf{A},\mathbf{Z}) + \log \left| \det \frac{\partial \mathbf{f}_\phi}{\partial \mathbf{Z_0}} \right|\right] 
\end{align*}

We call our jointly permutation-invariant generative model the \gls*{ge-vae}.



\section{Related Work}

\paragraph{Deep generative models of graphs.} So far the most successful inductive model of graph structure has been GraphRNN \citep{you_graphrnn:_2018}, which fits an auto-regressive model to sequences of node and edge formations derived from $\mathbf{A}$. The factorization implied by each sequence is dependent on chosen node orderings, so the model is not permutation invariant. However, by amortizing over sampled breadth-first orderings the model is approximately permutation invariant for modest graph sizes $|V|$. Graphite \citep{grover_graphite:_2019} is a variational auto-encoder model with permutation-equivariant MPNNs for encoding and decoding, but in the absence of node features relies on an arbitrary node ordering by setting node features $\mathbf{X}=\mathbf{I}$. \gls*{gnf} \citep{liu_graph_2019} uses permutation equivariant MPNNs to parameterize coupling layers in a normalizing flow model. They initialize node embeddings by sampling $\mathbf{X} \sim \mathrm{Normal}(\mathbf{0}, \sigma^2\mathbf{I})$, which is invariant to node re-orderings. However, they require a separate decoder for generating samples $\mathbf{A}|\mathbf{Z}$, by reverse message-passing. We note that without $\mathbf{X}$ both Graphite and \gls*{gnf} require message-passing over fully connected $\mathbf{A}$ to sample new graphs, a step which we replace with the Set Transformer.

\paragraph{Permutation invariant and equivariant models.} 

\citet{zaheer_deep_2017} first introduced permutation invariance and equivariance in the context of deep models. \citet{herzig_mapping_2018} introduced \emph{graph permutation-invariance} which is similar to our notion of joint permutation-equivariance under permutations of $\mathbf{A}$, but assumes the presence of unique node features $\mathbf{X}$ to break symmetry. \citet{hartford_deep_2018} discuss \emph{exchangeable matrix invariance}, which reflects separate symmetries in separate  permutations of rows and columns of $\mathbf{A}$, but not joint permutations. \citet{bloem-reddy_probabilistic_2019} provide a review of the above definitions that capture the symmetry in graphs. 


\section{Experiments}
\begin{figure}[t]
    \centering
    \includegraphics[width=0.9 \linewidth]{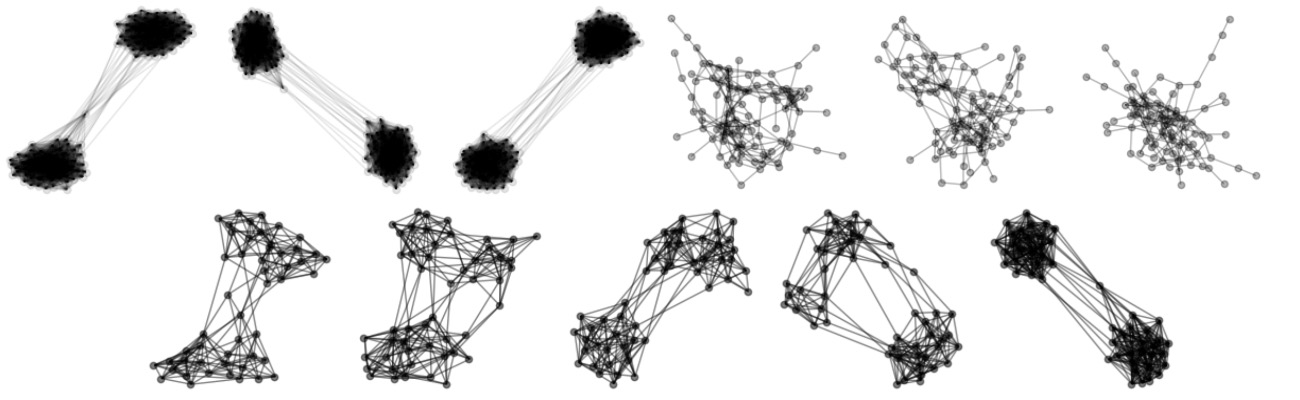}
    \caption{(Top) Generated graphs for the Community and Ego datasets. (Bottom) Interpolation in latent space between a four and two community graph structure.}
    \label{fig:generated}
\end{figure}

\vspace*{-5pt}
We experiment with several datasets, following the GraphRNN \citep{you_graphrnn:_2018} codebase. (1) Community: 3500 two-community graphs with ER clusters. (2) Ego: 757 3-hop ego networks extracted from Citeseer \citep{sen_collective_2008}. (3) Grid: 3500 standard 2D grid graphs. (4) Protein: 918 protein graphs over amino acids \citep{dobson_distinguishing_2003}. All datasets were split into roughly $\frac{2}{3}$ into training and $\frac{1}{3}$ into test sets. In order to handle graphs of various sizes in a dataset, we take only the eigenvectors corresponding to the smallest $P$ eigenvalues of each graph's unnormalized Laplacian. We implement masking in all self-attention steps, and maximize the reconstruction log-probability \emph{per edge} (i.e. per dimension). We evaluate by reporting \gls*{mmd} statistics over degree distributions, clustering coefficient distributions, and orbit count statistics (Table \ref{tab:table}). For the \gls*{ge-vae} we report estimated test set log-likelihoods as well. We exhibit visualizations of generated graphs in Figure \ref{fig:generated}. Overall we find that the \gls*{ge-vae} is competitive with GraphRNN on the Community and Ego datasets, but outperformed by GraphRNN on the Grid and Protein datasets. We suspect that this is due to the extremely multimodal nature of these graphs that are difficult to capture with a latent-variable model.

\begin{table}[t]
\centering 
\small
\begin{tabular}{lrrrrrrrrr}
\toprule
    &&& \multicolumn{4}{c}{Graph Embedding VAE} & \multicolumn{3}{c}{GraphRNN} \\ \cmidrule(l{2pt}r{2pt}){4-7} \cmidrule(l{2pt}r{2pt}){8-10}
       Dataset & $\max |V|$ & $\max |E|$&  bits/dim & degree &  cluster & orbit & degree & cluster & orbit \\
\midrule
 Community & 160 & 1945 & 0.297  & 0.011 & 0.056 &  0.002 & 0.014 & 0.002 & 0.039\\
           Ego & 399 & 1071 & 0.155  &  0.116 & 0.711 & 0.163 & 0.077 & 0.316 & 0.030\\
      Grid & 361 & 684 & 0.071 & 0.779 & 0.026 & 0.509 & $10^{-5}$ & 0 & $10^{-4}$\\
       Protein & 500 & 1575 & 0.114  & 0.591 & 1.563 & 0.451 & 0.034 & 0.935 & 0.217\\ \bottomrule\\
\end{tabular}
\caption{Test set MMD and log-likelihood graph generation statistics comparing our method \gls*{ge-vae} with GraphRNN. Results for GraphRNN are reported from \citet{you_graphrnn:_2018}.}
\label{tab:table}
\end{table}

\section{Discussion}
\begin{figure}[h] 
    \centering
    \includegraphics[width=\linewidth]{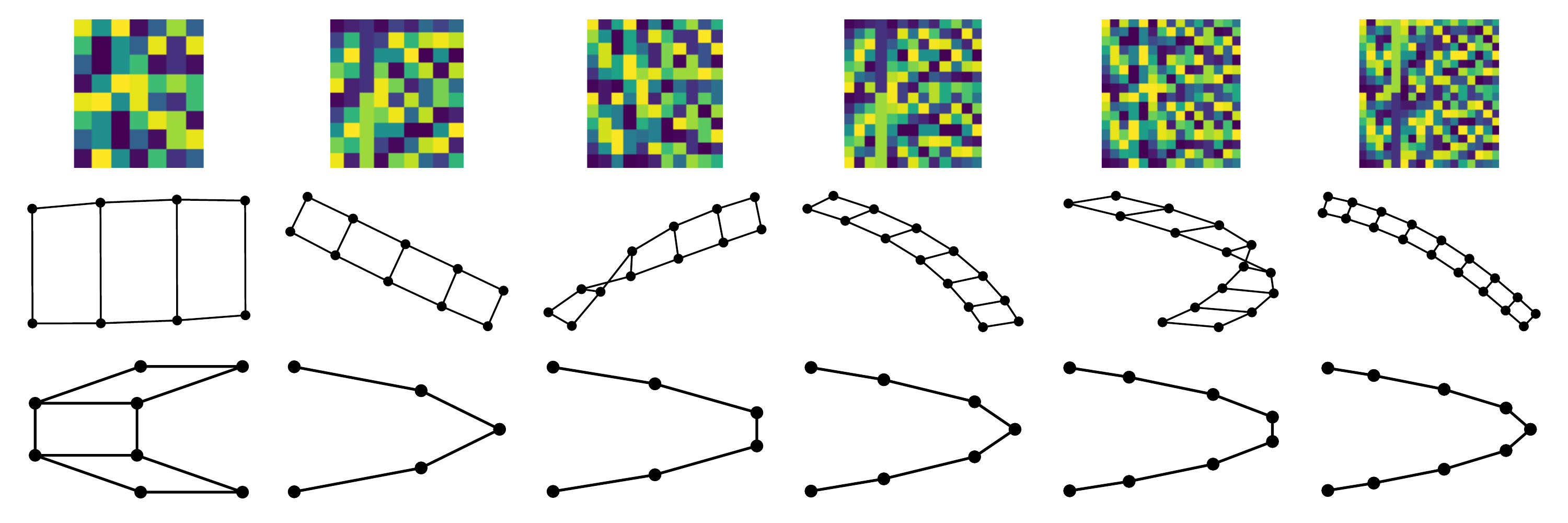}
    \caption{Laplacian Eigenmap embeddings for a sequence of ladder graphs of increasing size. The top row shows the embeddings and the bottom row shows the projections onto the first two columns (i.e. eigenvectors). As the size of the graph grows, the number of dimensions necessary to different betweeen the two columns of the ladder increases.}
    \label{fig:ladder_example}
\end{figure}

\vspace*{-5pt}
\paragraph{Embedding limitations.} The primary limitation of our method is heavy dependence on graph embeddings to encode the structure of the adjacency matrix in a jointly permutation-equivariant way. There is a heavy upfront computation cost to calculating embeddings for large graphs. Moreover, graph embeddings (such as the Laplacian Eigenmap) are not generally scale-invariant; as graph size increases, the representations encoded each dimension of the node embeddings do not straightforwardly map between graphs (see Figure \ref{fig:ladder_example} in the \nameref{sec:appendix}).
\vspace*{-5pt}
\paragraph{Meta-learning embeddings.} A meta-learned graph embedding model that is simultaneously trained along with our latent-variable model would allow for more flexible representations. Such a process would need to be permutation-equivariant, and also able to break symmetries by being position-aware (a simple \gls*{mpnn} would not suffice) \citep{you_position-aware_2019}. We leave this for future work.

\section*{Acknowledgement}
\vspace*{-10pt}
This work was supported by NRF-2019M3E5D4065965 project.

\newpage
\bibliographystyle{apalike}
\bibliography{neurips_2019.bib}

\newpage
\section{Appendix}

\label{sec:appendix}

\paragraph{Injective flow perspective.} Suppose we instead replace the first step of the encoder as
\begin{equation*}
    \cancel{\mathbf{Z}_0 | \mathbf{A} \sim \mathrm{Normal}(\mathrm{Embed}(\mathbf{A}),\sigma^2 \mathbf{I})}\quad\quad 
    \mathbf{Z}_0 | \mathbf{A} = \mathrm{Embed}(\mathbf{A}).
\end{equation*}
Then applying $\mathbf{f}_\phi \circ \mathrm{Embed}$ results in a series of transformations of $\mathbf{A}$ into the latent variable $\mathbf{Z}$. Crucially, graph embedding methods are not invertible, since there will always exist embeddings that do not correspond to any adjacency matrices -- so we cannot interpret the composition as a series of invertible flows. However, when the graph embedding is injective (such as the case when the Laplacian Eigenmap is used \citep{verma_hunt_2017}), the result is an injective flow \citep{kumar_learning_2019}.
\begin{align*}
    \log p_\phi(\mathbf{A}) & = \log p(\mathbf{f}_\phi \circ \mathrm{Embed} \circ \mathbf{A}) + \frac{1}{2}\log\left |\det J_{\mathbf{f}_\phi \circ \mathrm{Embed}}(\mathbf{A})^\top J_{\mathbf{f}_\phi \circ \mathrm{Embed}}(\mathbf{A}) \right|
\end{align*}

\paragraph{Log-likelihood evaluation.} We compute test set log-likelihood by Monte Carlo importance sampling with the variational posterior, using 128 samples. 
\begin{align*}
    \log p_\theta(\mathbf{A}) & = \log \int_\mathbf{Z} p_\theta(\mathbf{A},\mathbf{Z})d\mathbf{Z}\\
    & = \log \mathbb{E}_{q_\phi(\mathbf{Z}|\mathbf{A})}\left[ \frac{p_\theta(\mathbf{Z},\mathbf{A})}{q_\phi(\mathbf{Z}|\mathbf{A})}\right]\\
    & = \log \mathbb{E}_{q_\phi(\mathbf{Z}|\mathbf{A})}\left[ \frac{p(\mathbf{Z})p_\theta(\mathbf{A}|\mathbf{Z})\left|\det \frac{\partial \mathbf{f}_\phi}{\partial \mathbf{Z}_0} \right|}{q_\phi(\mathbf{Z}_0|\mathbf{A})}\right]
\end{align*}
We take the sum of the upper triangular entries of $p_\theta(\mathbf{A}|\mathbf{Z})$ and then divide by $\frac{1}{2}|V|(|V|-1)$ to calculate the number of bits per dimension, independent of the number of nodes in the graph.

\paragraph{Generating large-scale graphs}

We can generate large-scale graphs efficiently by interpreting the Bernoulli-Exponential link as augmented Poisson random variables instead of augmented Exponential random variables. Recalling that the marginal $p(a_{i,j}=1) = 1-e^{-\mathbf{z}_i^{\ast^\top} \mathbf{z}^\ast_j}$, let
\begin{equation*}
    a_{i,j} | \mathbf{z}^\ast_i, \mathbf{z}^\ast_j \sim \mathrm{Bernoulli}(m_{i,j} = 0) \quad\quad m_{i,j} | \mathbf{z}^\ast_i, \mathbf{z}^\ast_j \sim \mathrm{Poisson}(\mathbf{z}_i^{\ast^\top} \mathbf{z}^\ast_{j}).
\end{equation*}
It follows that the total number of edges is distributed as
\begin{align*}
    E = \sum_{i < j} a_{i,j} & \sim \mathrm{Poisson}\left(\sum_{i < j}\mathbf{z}_i^{\ast^\top}\mathbf{z}_j^\ast\right)\\
    & \sim \mathrm{Poisson}\left(\frac{1}{2} \left( (\sum_{i=1}^n \mathbf{z}^\ast_i)^\top (\sum_{i=1}^n \mathbf{z}^\ast_i)- \sum_{i=1}^n ||\mathbf{z}^\ast_i||^2_2 \right)\right)\\
    & \sim \sum_{d=1}^{P} \mathrm{Poisson}\left(\frac{1}{2}(\sum_{i=1}^n z^\ast_{i,d})(\sum_{i=1}^n z^\ast_{i,d})- \sum_{i=1}^n z^{\ast^2}_{i,d} \right)\\
    & \overset{\mathrm{def}}{=} \sum_{d=1}^P E_d.
\end{align*}
So we can first sample the total number of edges $E$ by sampling $E_d$ (corresponding to each dimension), then sample the nodes $(i,j)$ corresponding to each edge by picking
\begin{equation*}
  p(i) | E_d \propto z^\ast_{i,d}.
\end{equation*}

\newpage
\textbf{Additional figures.}

\end{document}